\documentclass{article}
\usepackage{graphicx} 
\usepackage[margin=1in]{geometry}
\RequirePackage{doi}
\usepackage{hyperref}
\usepackage{setspace}   
\usepackage{natbib}
\usepackage[utf8]{inputenc}
\usepackage[T1]{fontenc}
\usepackage{xcolor}
\usepackage{tcolorbox}
\usepackage{tikz}
\usepackage{booktabs}
\usepackage{multirow}
\usepackage{siunitx}
\usepackage{longtable}
\usepackage{subfigure}
\usepackage{float}
\usepackage{caption}
\usepackage{subcaption}

\definecolor{lightblue}{RGB}{173,216,230}
\definecolor{lightgreen}{RGB}{144,238,144}
\definecolor{lightorange}{RGB}{255,229,180}
\definecolor{lightred}{RGB}{255,182,193}
\definecolor{lightpurple}{RGB}{230,190,255}

\newenvironment{example}
{\begin{tcolorbox}[colback=lightblue!10,colframe=lightblue!50!black,title=Example]}
{\end{tcolorbox}}

\newcommand{\colortext}[2]{\textcolor{#1}{#2}}

\usepackage{soul}

\title{ConfliBERT: A Language Model for Political Conflict\footnote{Previous versions have been presented at the virtual Methods in Event Detection Colloquium (September 2024). This research was supported by NSF award 2311142;  
used Delta at NCSA / University of Illinois through allocation CIS220162 from the Advanced Cyberinfrastructure Coordination Ecosystem: Services \& Support (ACCESS) program, which is supported by NSF grants 2138259, 2138286, 2138307, 2137603, and 2138296. This material is based upon High Performance Computing (HPC) resources supported by the University of Arizona TRIF, UITS, and Research, Innovation, and Impact (RII).  Any opinions, findings, and conclusions or recommendations expressed in this material are those of the author(s) and do not necessarily reflect the views of the National Science Foundation,  NCSA, or the affiliated university resource providers. Thanks for suggestions go to Scott Althaus, R. Michael Alvarez, Ben Bagozzi, Rebecca Cordell, Ryan Kennedy, Hyein Kim, Shahryar Minhas, Philip Schrodt, and Nora Webb Williams.}}

\author{Patrick T.~Brandt; UT Dallas, pbrandt@utdallas.edu\\
Sultan Alsarra; King Saud University, salsarra@ksu.edu.sa\\
Vito J.~D’Orazio; West Virginia University, vito.dorazio@mail.wvu.edu\\
Dagmar Heintze; UT Dallas, dagmar.heintze@utdallas.edu\\
Latifur Khan; UT Dallas, lkhan@utdallas.edu\\
Shreyas Meher; UT Dallas, shreyas.meher@utdallas.edu\\
Javier Osorio; University of Arizona, josorio1@arizona.edu\\
Marcus Sianan; UT Dallas, marcus.sianan@utdallas.edu}

\date{\today}

\begin{document}

\maketitle

\begin{abstract}
    Conflict scholars have used rule-based approaches to extract information about political violence from news reports and texts. Recent Natural Language Processing developments move beyond rigid rule-based approaches. We review our recent ConfliBERT language model (Hu et al.~2022) to process political and violence related texts. The model can be used to extract actor and action classifications from texts about political conflict.  When fine-tuned, results show that ConfliBERT has superior performance in accuracy, precision and recall over other large language models (LLM) like Google's Gemma 2 (9B), Meta's Llama 3.1 (7B), and Alibaba's Qwen 2.5 (14B) within its relevant domains. It is also hundreds of times faster than these more generalist LLMs.  These results are illustrated using textsfrom the BBC, re3d, and the Global Terrorism Dataset (GTD).  
\end{abstract}

\newpage 

\section{Introduction}
\doublespacing

A common text as data problem in the social sciences is the need to migrate information in a large corpora into a data summary of that text.  As social (political) scientists we possess replicable and encoded domain expertise to understand these texts.  There are components of the \emph{processing} and \emph{summarization} of large corpora that are better understood by experts in machine learning and natural language processing (NLP).  How then should one combine the insights of domain experts and computational and information scientists of natural language?   Here we present one possible answer using recent large language models to illustrate how domain expertise and computational science can be combined. 

The conflict event data we care about are typically from human-coded news reports. The transformation of unstructured news texts into structured "who-did-what-to-whom" event data is fundamental in the international relations, and studies of conflict and political violence. The process of gathering and preparing data for analysis in this domain is: assemble a corpus, filter for relevant information, identify target events, and annotate event attributes. This process can be costly and time consuming: time required for data collection, filtering the large amount of unstructured text, training human annotators on the usage of the applied ontology, and several rounds of quality controls to ultimately achieve a curated text corpus. This parallels the widely systematized way to approach text-as-data in much of international relations and the social sciences \citep{o2013learning,grimmer2022text}. Ultimately, these pre-processing steps are applied to the unstructured corpus to permit the extraction of relevant information on actors (who), actions (did what) and the targets (to whom) for political science research.   

This transformation of unstructured news text faces several well-known challenges that researchers in the field have long faced.\footnote{We presume the existence of a relevant corpus of texts in digital form, akin to economists' can-opener assumption.} These can be categorized into three main areas:

\begin{enumerate}
\item Filtering for relevant information or reports
\item Identifying events
\item Annotating event attributes 
\end{enumerate}

While these areas can be challenging, they align with common tasks in language learning and computational information science. In NLP these latter tasks are addressed through techniques such as semantic role learning (SRL), automated content extraction (ACE). Here we address how these problems can be addressed with a language model called ConfliBERT or a Bidirectional Encoder Representations from Transformers (BERT) model trained on conflict data.  ConfliBERT, as a political ``conflict'' informed model, does this using a BERT model.\footnote{The conflict domain part is first and the LLM / AI tool is second. Since we started this in 2021, the LLM  tool of the time was BERT.  However, we further introduce a conflict-specific adaptation of LLAMA \citep{dubey2024llama}, a generative AI-based model, which we termed ConfLlama and trained on conflict-specific training data in section 4 to permit a performance comparison between ConfliBERT and current generative AI models.}  ConfliBERT is a language model trained on domain-specific conflict and political violence data and achieves superior performance in the summarization and classification tasks political scientists perform on unstructured text corpora.  ConfliBERT (for English) is publicly available at Hugging Face, and is trained on an expert-curated corpus of 33.7 GB about conflict and political violence.  Combining domain-specific knowledge and computational tools in ConfliBERT improves unstructured conflict texts processing.  The method discussed here is a domain-specific pretrained language model implemented for English \citep{hu2022conflibert}, and there are variants for Spanish \citep{yang2023conflibert}, and Arabic \citep{alsarra2023conflibert}.   ConfliBERT, as a political science trained model, can help encode information about events from text without having to have a fully specified ontology of actors or their interactions (a major impediment in earlier domain-based methods).


Our proposed method uses a transformer language models \citep{devlin2018bert} and large amounts of politically relevant news texts as training data.   We re-cast the problems of the political science domain into those more commonly seen in the information and computer science domains of text and natural language processing and inferences.  This trades human annotation and classification costs for computational resources (which grow more powerful and cheaper), while augmenting the speed of data generation. However, how this is done in the computational linguistics and information sciences community needs to be bridged to how social scientists think about these problems.  There the focus is on annotation of spans of text that correspond to linguistic and contextual entities (by human or machine).  In contrast, political and social scientists focus on event attributes, their modality and characteristics  \citep{olsen-etal-2024-socio}.  Similar domain-specific BERT models have shown to equally outperform generic BERT models in text classifications in other scientific fields such as: biomedical SCIBERT \citep{beltagy2019scibert}, material sciences, MatSCIBERT \citep{GuptaTanishq2022MAmd}, legal LegalBERT \cite{Legal_BERT}, finance (FinBERT) \cite{araci2019finbertfinancialsentimentanalysis}, clinical notes (ClinicalBERT) \citep{huang2020clinicalbertmodelingclinicalnotes}, and patent texts (patentBERT) \citep{lee2019patentbertpatentclassificationfinetuning}. 


For the political science domain we want to use ConfliBERT to accomplish three key information extraction and summarization tasks that are part of "coding event data":

\noindent \textbf{Filtering for Relevant Information}: In political science, the sheer volume of available text data --- from news articles and social media posts to diplomatic cables and policy documents --- necessitates efficient filtering. The challenge lies not just in identifying texts that contain political content, but in discerning which texts contain information relevant to specific research questions or event types of interest.

For instance, a researcher must differentiate between reports of actual violent events and discussions of potential conflicts or historical references. This task becomes even more complex when dealing with subtle forms of political interaction or when trying to capture early warning signs of emerging conflicts. ConfliBERT's binary classification for political violence addresses this challenge by providing a confidence score for the relevance of a given text to conflict or political violence. This allows researchers to quickly sift through large volumes of text, focusing their attention on the most pertinent sources.

\noindent \textbf{Identifying Events:} Once relevant texts are identified, the next challenge is to accurately identify and delineate specific events within these texts. This task is complicated by several factors. Multiple events may be described within a single text, or events may be described across multiple sentences or even multiple documents. The language used to describe events can be ambiguous or metaphorical, and different sources may provide conflicting accounts of the same event.

In political science, accurate event identification is crucial for creating reliable event datasets, which form the backbone of many quantitative analyses in the field. Misidentification of events can lead to skewed results and flawed policy recommendations. ConfliBERT's multiclass classification contributes to this task by categorizing the type of event described in the text (e.g., Armed Assault, Bombing or Explosion). This classification can serve as a starting point for more detailed event extraction and coding.  But this step often needs iterations and revisions thus requiring speed and computational efficiency as well as accuracy.

\noindent \textbf{Annotating Event Attributes:} Perhaps the most challenging aspect in unstructured text processing is the detailed annotation of event attributes—the "who," "what," "to whom," "where," and "when" of each identified event. This requires not just entity recognition, but also understanding the roles these entities play in the event and the relationships between them.

Political scientists often need to code additional attributes beyond the basic event structure. These might include the intensity or scale of the event, the motivations or goals of the actors involved, the immediate and long-term consequences of the event, and the broader political context in which the event occurred. ConfliBERT's Named Entity Recognition (NER) capabilities provide a foundation for this task by identifying key actors, organizations, and locations within the text. However, fully automating the annotation of complex event attributes remains a significant challenge and often requires a combination of machine learning techniques and expert human coding.

We compare our model to some of the latest generative large language models and show its highly favorable performance on the tasks defined above.  ConfliBERT significantly alleviates the human annotation bottleneck in these three areas.  We also review the use of ConfliBERT and our related innovations for engineering data from text. ConfliBERT was trained on conflict and political violence related data in 2021, and consistently outperforms original BERT \citep{hu2022conflibert}, Electra and RoBERTa \citep{osorio_2024}.  Its non-English variants, ConfliBERT-Spanish outperforms BERT variants like mBERT and BETO \cite{yang2023conflibert}; and, ConfliBERT-Arabic does the same relative to AraBERT \citep{osorio_2024}.

\emph{Furthermore, ConfliBERT outperforms current generative AI models such as Meta's Llama 3.1 \citep{dubey2024llama}, and Gemma 2 \citep{team2024gemma} in binary classification and NER tasks.}  Here we compare ConfliBERT to nascent generative LLMs like Google's Gemma 2, Meta's Llama, and Alibaba's Qwen 2.5 \citep{hui2024qwen25codertechnicalreport}.  We demonstrate that the fine-tuned models used can outperform the more general models in terms of accuracy, precision, recall, and computational time.  

ConfliBERT has several advantages over comparable contemporaneous methods for machine coding events.  First, it is easily deployed and replicable as a method since it is open source and can be deployed on conventional hardware.  Second, it is significantly better on comparable, relevant quality metrics and faster than rival or even newer generative artificial intelligence (AI) methods that used encoder-decoder technologies with graphical processing units (GPUs). Third, it can be rapidly deployed to detect new event data and their characteristics.\footnote{We thank a reviewer / commentators for emphasis on this point.}  This means it can be tuned and adjusted as needed for new cases, data, and texts.  This allows users to improve the extraction, coverage (geographically, and as we show, linguistically), across new data and training domains.    Fourth this means that additional downstream tasks such as recoding texts, extracting additional variables or features, etc. are all much faster and easier than what has historically been the case.  We show this in our examples where differences are seen in the classifications of terrorist event types in the GTD dataset across the LLMs.  ConfliBERT and domain-specific models provide much better results compared to generalist LLMs like Gemma, Llama, and Qwen.

\section{The Event Coding Problem}

In event data research, scholars break down texts into key attributes: actors (sources and targets), actions, locations, and dates.\footnote{For those interested in getting event data and not exploring the weeds of this process, see \citet{halterman2023plover}'s Political Language Ontology for Verifiable Event Records (PLOVER) and the POLitical Event Classification, Attributes, and Types (POLECAT) dataset. which are a record of domestic and international political interactions described in international news reports from 2010 to the present. The news reports are in English or machine translated into English from Arabic, Chinese, French, Portuguese, Russian, or Spanish before they are coded.}  With actor coding, there are two broad approaches that eschew human coding: mining past data to propose new groups or categories of actors \citep{8258064} and machine learning or transformer approaches utilizing BERT-based models \citep{dai2022political,hu2022conflibert,skorupa2022multi, alsarra2023conflibert,yang2023conflibert,halterman2023creatingcustomeventdata}. To code actions prior work employed sparse parsing using human-annotated dictionaries \citep{schrodt2001automated, osorio2020supervised}, whereas newer approaches handle new ontology or action extensions through either up-sampling \citep{halterman2021few}, natural language inference (NLI) \citep{dai2022political,hu2022conflibert,lefebvre2022rethinking,skorupa2022multi,halterman2023plover,croicu2024deepactivelearningdata}, or zero-shot prompts \citep{hu-etal-2024-leveraging}. Geographic coding in earlier work relied on the location inferred from the actors to identify where the event occurred. Some approaches to determine location use sparse parsing \cite{osorio2020supervised}, word embedding and NER \citep[e.g.,]{halterman2017mordecai,imani2017focus,imani2019did}, and even BERT \citep{halterman2023plover}.  For date or time coding of events, researchers generally parse the byline of the news report to acquire the publication date \citep{osorio2020supervised}, but the publication and the event occurrence dates are not always the same. A recent approach is to apply BERT technology to reconcile date information from the news story \citep{halterman2023plover}.  All of these information extraction approaches are prone to various errors \citep{brandtsianan2024}, and latest methods attempt to reduce them using BERT-alike language models. 

ConfliBERT provides a domain level solution to a common problem.  For most generic generative and extractive artificial intelligence (AI) tasks, an LLM needs broad training.  These training steps generate huge costs in terms of 1) training data, 2) human / expert time, and 3) computational complexity to combine and produce the relevant model. In a domain-specific application like ConfliBERT, several choices make these challenges much more feasible for a social science tool.  First, creating an extractive LLM or a BERT LLM (or even for that matter a \emph{simple} predictive or generative suggestion model) can be done much more rapidly and cheaply.  Since there is domain knowledge and insight provided in the initial training steps, steps 1 and 2 above for training a generic LLM are greatly scaled back, resulting in a superior model in a shorter period of time.  Second, ConfliBERT can then be augmented or expanded in various ways (which we demonstrate below) to focus on harder tasks others have attempted, such as ontology extension \citep{radford2021automated}, actor detection and recognition \citep{apart,8258064}, and image processing applications \citep{Threlkeld_2019, wen-etal-2021-resin}.



The earlier needs to extract actors, action events (verbs) and additional information from texts for political science and international relations studies of conflict are accommodated in three different NLP tasks.  The three main tasks that ConfliBERT addresses are:

\begin{description}
    \item [Classification]  Which texts contain relevant information about politics, conflict, violence?  We give examples of this below based on data from the BBC and re3d text corpora.   These are
    \begin{enumerate}
        \item binary classifications: yes / no questions
        \item multi-label classifications: in a series of reports about protests, which types of protest are present (labor, peaceful, violent, etc.)?
    \end{enumerate}
    \item [Named Entity Recognition (NER)]  What are the ``who'' and ``whom'' that characterize the event?  These are most typically the linguistic subjects and objects of the sentences and clauses (subject to textual disambiguation and co-referencing).  But making sense of them becomes a task for a political scientist to identify the source / initiator of a political event toward a target or other political actor.  We give an example below of doing this for texts about terrorist attacks from GTD.
    \item [Masking / Coding new entities and / or events] is the extension of any ontology of new kinds of events.  This can include teaching a model which events are new ones, ones to be excluded, or newly emergent actors and their roles.  
\end{description}

The first two of these tasks are \emph{extractive} tasks often best handled using statistical or machine learning algorithms to make suggestions for validation.  These tasks are predictive in the sense that a model is trained to learn and predict patterns based on repeated past examples or interactions.  Prior SVM-like versions of similar classification problems for international conflict events have been addressed with simpler models that are context free \citep{DOrazio_Landis_Palmer_Schrodt_2014}.   But they are the context-dependent tasks for which BERT-alike models are optimized to do.   A method like ConfliBERT improves these previous approaches using longer embedded patterns of related text and their rapid abilities to find additional patterns faster than a human.  It is able to accomplish this in situations where the events or entities to be classified are rare (there may be few examples to mask or learn from) or where there is a class imbalance (i.e., the presence of mass atrocities is rare, even among a collection of stories about multiple crimes and political attacks).

The last task is harder, but can begin with an LLM or a BERT-like model.  Determining whether another event is like a prior one (or a related class or actor) requires providing examples that omit the thing to be predicted (masking or hiding it) and then seeing how well the model does.  This problem can apply to new actors and events and the determination of whether the model / coder is correct relies on the domain knowledge of the social scientist.  Further, the identification of new actors is often a masking task for which BERT-alike models are designed.\footnote{Our recent work in actor detection proposes Distant Supervised (DS) and Zero-Shot (ZS) approaches for extracting political actors and their roles using pre-trained language models \citep{parolin2021come,hu-etal-2024-leveraging}.} 

More generally these are problems of open information extraction (OIE), a critical NLP task. OIE extracts structured information from unstructured source text by detecting relationships without relying on a previously defined onthology.  These are generic text and information extraction tasks that are necessary in any coding of texts as data or processing human produced texts.  For a political scientist there are \emph{domain specific versions} of these tasks.   In the actor identification domain, OIE is a necessary task for accurate event detection and action feature classification.  

The problem of finding political actors is twofold. One challenge lies in the ability to automatically extract new political entities from text. The other is to correctly associate these actors with their respective political environment (role, country, organization, and group, etc.). But note this is \emph{not} specific to the domain of political science (remove the word ``political" from the first sentences of this paragraph).   To place an event in context, it is often critical to know not only an actor and their role, but also the actor’s relation to the government (e.g., a rebel group) or other organizations or individuals.  Across conflict research, actors are a fundamental unit of analysis: individuals, groups, organizations, countries, etc. Actors have attributes, such as a role within government or an affiliation with a particular group, and the attributes of actors can change over time. Accurately extracting conflict events from text, or building networks of political interactions, depends on our ability to identify actors and their roles. For example, a meeting between Joe Biden and Emmanuel Macron, two heads of state, is categorically different from a meeting between two elected officials from the same country.  Actor identification is a domain knowledge encoding of information into labels that are relevant to the political scientist. It requires the expertise of the political science scholars.
Typically, actor dictionaries associate individuals with their groups and actions \citep{schrodt2008cameo,boschee2015icews}.  The high cost and human effort to update these repositories is unsustainable and requires continual updating. This is particularly problematic when an actor’s role changes or a new actor emerges.  All of these complexities add to the costs and domain knowledge necessary for this component of extracting event data information from corpora.

\section{ConfliBERT Details and Examples}

ConfliBERT is a LLM trained on a \emph{curated corpus of high-quality text data about politics, conflict, and violent events} from newswires and expert sources in English.\footnote{Training examples are listed at \url{https://github.com/eventdata/ConfliBERT/tree/main/pretrain-corpora} and test or evaluation ones are at \url{https://github.com/eventdata/ConfliBERT/tree/main/data}.}  ConfliBERT uses this domain knowledge with which it has been ``trained'' to be a more useful language model than vanilla BERT, LLM, or a simple dictionary approach.  ConfliBERT is an engine or baseline for extracting information about political texts.  It 1) sorts political violence texts from other ones (classification), 2) identifies possible political actors and entities (since it was trained to do so and does NER with this knowledge), and 3) provides for masking and question and answer (QA) tasks for coding. Clearly, the process and methodology here can be adapted to use other methods beyond BERT (other options and more recent LLMs are explored below).   It can then be extended in domain areas / expertise, as well as scope conditions to include new languages, etc. via masking, fine-tuning, or other extensions.

ConfliBERT enables three distinct yet interconnected tasks that are crucial for a nuanced understanding of conflict dynamics: binary classification of violence-related news, multi-class classification of attack types, and named entity recognition in terrorism event reports.  Users then take this model as given and a base against which improvements can be measured.\footnote{The base ConfliBERT model is documented at \url{https://eventdata.utdallas.edu/} and \url{https://github.com/eventdata/ConfliBERT-Manual}.  Those wishing to deploy the model can access it directly via \url{https://huggingface.co/eventdata-utd}.  A user interface to evaluate short paragraphs and sentences is also available: \url{https://eventdata.utdallas.edu/conflibert-gui/} and \url{https://huggingface.co/spaces/eventdata-utd/ConfliBERT-Demo}.
We then fine-tune the model simultaneously on all three tasks, using a weighted loss function that balances the importance of each task. To enhance the model's performance on individual tasks, we implement task-specific data augmentation techniques. }

An example of the first task is the binary classification of news articles to determine their relevance to gun violence. Utilizing a dataset comprising BBC news articles and the 20 Newsgroups corpus, we trained ConfliBERT to discern whether a given news item pertains to gun violence incidents. This fine-tuning task is significant for both domestic and international conflict studies, since it shows how to filter rapidly large volumes of news data about something like gun violence-related events. The ability to quickly identify relevant articles from a diverse news corpus can significantly enhance researchers' capacity to track and analyze (gun-related) conflicts in real-time.

For this binary classification task, ConfliBERT distinguishes between \emph{gun violence-related and non-gun violence-related incidents.} Consider these examples:

\begin{example}
\textcolor{blue}{Input:} "Two Lashkar e Jhangvi LeJ militants Asim alias Kapri and Ishaq alias Bobby confessed to killing four Rangers in Ittehad Town of Karachi, the provincial capital of Sindh."\\
\textcolor{green}{Output:} Gun Violence Related (1)

\textcolor{blue}{Input:} "More than a week after a woman Communist Party of India-Maoist (CPI-Maoist) cadre was killed in an encounter in the forests of Lanjigarh block in Kalahandi District, the Maoists identified her as Sangita and called a bandh (general shutdown) in two Districts in protest against the killing."
\textcolor{green}{Output:} Gun Violence Related (1)
\end{example}

The second task expands on this binary classification to a more nuanced multi-class classification of attack types. Employing the Global Terrorism Database (GTD) to train ConfliBERT, it can classify attacks into nine distinct categories, including bombing/explosion, armed assault, assassination, and various forms of hostage-taking. Here are examples from the South Asia Terrorism Portal (SATP) dataset \href{https://satp.org/}:

\begin{example}
\textcolor{blue}{Input:} "Islamic State (IS) in the latest issue of its online magazine Dabiq claimed that the five of the nine Gulshan café attackers were suicide fighters... The mujahidin held a number of hostages as they engaged in a gun battle with apostate Bengali police and succeeded in killing and injuring dozens of disbelievers before attaining shahadah."\\
\textcolor{green}{Output:} Armed Assault

\textcolor{blue}{Input:} "The ongoing construction work of an interstate bridge on Pranhita River on Maharashtra-Telangana border was thwarted by the Naxalites [Left Wing Extremists, LWEs] who set an excavator on fire and also damaged other equipment at the construction site at Gudem in Aheri taluka (revenue unit) of Gadchiroli District on April 26."\\
\textcolor{green}{Output:} Facility/Infrastructure Attack

\textcolor{blue}{Input:} "Three boys sustained injuries when a landmine went off in Atmar Khel area of Baizai tehsil (revenue unit) in Mohmand Agency of Federally Administered Tribal Areas (FATA) on June 18."\\
\textcolor{green}{Output:} Bombing/Explosion
\end{example}

The third task ConfliBERT addresses is named entity recognition, crucial for extracting structured information from unstructured text, enabling more detailed and systematic analyses of conflict actors and targets.  Using event reports (MUC-4), which contain annotations of terrorism events, we train ConfliBERT to identify and classify entities such as Perpetrator Organizations, Physical Targets, Victims, and Perpetrator Individuals.  Here is an NER classification example using SATP:

\begin{example}
\colortext{blue}{Input:} "A senior \colortext{orange}{Muttahida Qaumi Movement (MQM)} \colortext{orange}{[ORG]} worker identified as \colortext{red}{Sohail Rasheed} \colortext{red}{[PERSON]}, 30, was shot dead near his home in \colortext{purple}{Naeemabad} \colortext{purple}{[LOC]} in \colortext{purple}{Korangi Town} \colortext{purple}{[LOC]} of \colortext{purple}{Karachi} \colortext{purple}{[LOC]}, the provincial capital of \colortext{purple}{Sindh} \colortext{purple}{[LOC]}, on \colortext{green}{June 19} \colortext{green}{[DATE]}."\\
\colortext{blue}{Output:}\\
\colortext{orange}{Perpetrator Organization:} Muttahida Qaumi Movement (MQM)\\
\colortext{red}{Victim:} Sohail Rasheed
Physical Target: Not specified\\
\colortext{purple}{Location:} Naeemabad, Korangi Town, Karachi, Sindh\\
\colortext{green}{Date:} June 19
\end{example}

The versatility demonstrated across these tasks also suggests potential applications in related fields such as international relations, security studies, and public policy. By providing a tool that can simultaneously categorize events, identify key actors and targets, and filter relevant information from large text corpora, ConfliBERT offers a powerful means of analyzing the complex landscape of modern conflicts.

Recall, that ConfliBERT was trained in 2021 on data that at this point is nearly 3 years old \citep{hu2022conflibert}.  So a question is, how well does it do with more contemporaneous events and data?  Consider the following example that has been processed using the interface at \url{https://eventdata.utdallas.edu/conflibert-gui/} or \url{https://huggingface.co/spaces/eventdata-utd/ConfliBERT-Demo}:

\begin{quote}
\singlespacing
Former President Donald Trump, the 2024 presumptive Republican presidential nominee, was escorted off the stage by Secret Service after gunshots were fired at his rally in Butler, Pennsylvania. Mr.~Trump was injured from the incident, with blood appearing on the right side of his face. This occurred two days before the start of the Republican National Convention in Milwaukee. The Butler County, Pennsylvania, district attorney told the Associated Press that a shooter was dead and a rally attendee was killed.\footnote{\href{https://www.c-span.org/video/?536813-1/president-donald-trump-removed-stage-shots-fired-pennsylvania-rally}{``Former President Donald Trump Removed From Stage After Shots Fired at Pennsylvania Rally'', CSPAN, July 13, 2024. Accessed 2024-09-09.}
}
\end{quote}

\noindent The outputs for each of the coding tasks are:
\begin{description}
    \item[Binary Classification for Political Violence]  "Positive: The text is related to conflict, violence, or politics. (Confidence: 99.85\%)"
    \item[NER] entries returned are:
    \begin{description}
    \item[Location: ] ``butler , pennsylvania'', ``milwaukee'', ``butler county , pennsylvania''
    \item[Organisation: ] ``secret service'', ``republican national convention'', ``the associated press'', ``district''
    \item[Person: ] ``former president donald trump , the 2024 presumptive republican presidential nominee'', ``attorney''
    \end{description}
\item[Multilabel Classification]  ``Armed Assault (Confidence: 98.40\%) \/ Bombing or Explosion (Confidence: 5.39\%) \/ Kidnapping (Confidence: 0.44\%) \/ Other (Confidence: 0.95\%)''.
\end{description}

\noindent The only notable error is the inability to recognize ``district attorney'' as a person.  Such an error can easily be corrected with additional fine tuning about legal actors and titles.

The implications of this multi-faceted approach to conflict analysis using ConfliBERT are far-reaching for political science research. By addressing multiple aspects of conflict analysis simultaneously, researchers can process diverse datasets more efficiently, from broad news corpora to specialized conflict databases. 
The scalability of this approach is particularly noteworthy. Once trained on these diverse tasks, ConfliBERT can be rapidly deployed to process large volumes of new data, enabling real-time or near-real-time analysis of emerging conflicts. This capability is invaluable for researchers and policymakers who need to quickly assess and respond to evolving conflict situations.

\section{Evaluating ConfliBERT vs.~other LLMs}
\label{sec:eval}

\citet{hu2022conflibert} established ConfliBERT's superiority against a baseline of BERT \citep{devlin2018bert}.  They show that for binary classification and NER (the main tasks discussed here) ConfliBERT was better than BERT (based on using cased and uncased models) using $F_1$ and macro $F_1$ statistics for weighted precision and recall.  In prior analyses, it scored better than BERT uncased and cased models, in multiple tests and across datasets of interest to political conflict scholars (general news reports, gun violence reports, terrorist attack accounts [by region], etc.)\footnote{The diversity of comparisons include the BBC News dataset \citep{greene2006practical}, a sample SATP dataset \href{https://satp.org/}, the 20 Newsgroups dataset \citep{LANG1995331}, the Gun Violence dataset \citep{pavlick-etal-2016-gun}, and the Event Status dataset \citep{huang-etal-2016-distinguishing} for binary classification tasks. For NER tasks, ConfliBERT CONT cased achieved the highest macro F1 score for the source and target labeling NER task on the CAMEO Codebook \citep{gerner_2002} dataset, while ConfliBERT SCR uncased showed the highest macro F1 scores on both the MUC-4 \citep{muc-1992-message} and the \citep{re3d_2018} datasets.} \citet[see , Table 3 and Figure 1]{hu2022conflibert} \citet{haffner2023introducing} and \citet{croicu2024deepactivelearningdata} give additional and independent evidence of the model's strong performance relative to known alternatives for different conflict texts and related tasks.

The focus here is on ConfliBERT's efficacy in the two critical NLP tasks of binary classification (BC) and named entity recognition (NER) \emph{compared to recent developments like generative AI LLMs}.  Gauging ConfliBERT's comprehension and extraction of information from conflict-related texts can be benchmarked against more recently created baselines from much large LLMs like Gemma 2, LLama 3.1, and Qwen 2.5.  The goal is to assess the quality of an LLM like ConfliBERT and compare it to larger, more costly, and more computationally expensive alternatives.

We do this initially for two datasets that were used in the earlier comparisons of ConfliBERT to BERT: the BBC News Dataset and re3d.\footnote{Both datasets are available for public use and can be accessed through the ConfliBERT GitHub repository. 
}   The BBC News dataset is used for the binary classification task \citep{greene2006practical} and consists of 2,225 news articles, with 1,490 records for training and 735 for testing. The articles cover five categories: business, entertainment, politics, sport, and tech. For the conflict classification task, the dataset articles are relabeled as either conflict-related (1) or not conflict-related (0) by expert coders who analyzed each article's content and context. This dataset is particularly relevant providing a diverse range of news articles, thus testing ConfliBERT's performance in sorting conflict-related content across various domains. The binary classification task mimics real-world scenarios where analysts must quickly identify political conflict-relevant information from a stream of news articles.

Once such articles or reports are identified via binary classification, then political actor and action classification are the relevant NER tasks.  To compare ConfliBERT and more recent alternatives on this task, the re3d is used (Relationship and Entity Extraction Evaluation Dataset \url{https://github.com/dstl/re3d/}). These data are specifically designed for defense and security intelligence analysis, focused on the conflict in Syria and Iraq, providing domain-specific content across various source and document types with differing entity densities. The entities of interest include organizations, persons, locations, and temporal expressions. Ground truth labels were established by annotators using a hybrid process.\footnote{Described at \url{https://github.com/dstl/re3d/}.  A modified version of the re3d, which underwent several preprocessing steps is used:  These include tokenization and minor cleaning, removal of entity labels with confidence scores below 0.5, and resolution of overlapping entities by keeping only the largest span. The dataset was then converted to CoNLL 2003 format for compatibility with standard NER evaluation tools.}  The re3d is valuable to evaluate ConfliBERT's and LLMs' extraction of relevant entities from conflict-related texts.

\subsection{Methodology}

Across these two datasets in this section, the performance of a task is done using versions of 1) ConfliBERT, 2) Meta's Llama 3.1 (8B), 3) Google's Gemma 2 (9B), and 4) ConflLlama (8B). Note these are the most recent versions of these LLMs in late 2024.  (We leave the comparisons to the much larger Qwen model to the next section).  For the more recent LLMs, we offer brief descriptions:

\begin{description}
    \item[Meta's Llama 3.1] is the latest version of the Llama series of language models \citep{dubey2024llama}. With 7 billion parameters, it strikes a balance between computational efficiency and performance. 
    \item[Google's Gemma 2] has 9 billion parameters, represents a significant advancement in the field of large language models \citep{team2024gemma}, offering robust performance across a wide range of NLP tasks while maintaining a relatively compact size. 
    \item[Alibaba's Qwen 2.5] has a large pre-training corpus focused on math and coding. Another key improvement, especially in the context that we are using the model for is the greater accuracy in generating structure outputs (as JSON objects). 
    \item[ConfLlama] based on LLaMA-3 8B, was specifically fine-tuned on the Global Terrorism Database (GTD) using QLoRA with a learning rate of 2e-4 and LoRA rank of 8. The model was trained with gradient checkpointing enabled and 4-bit quantization, achieving convergence with loss reduction from 1.95 to approximately 0.90 \citep{conflllama}.  We employ both $Q4_{K_{M}}$ and $Q8_{0}$ quantizations for comprehensive performance analysis.
\end{description}


Various performance metrics quantify how well the models classify an event or its key attributes (e.g., actors, actions, locations, and dates). These metrics essentially compare the ground truth with what the machine extracts to produce a numerical result. This \textit{distance} between the two demonstrates the degree of congruence, and the goal for event data scientists is to achieve 100\% congruence across multiple possible sources of error \citep{brandtsianan2024,althaus2022total}.  For binary classification the precision, recall, and the \textit{F}\textsubscript{1} score are reported. 
The focus here is the $F_1$ statistic, the geometric mean of the precision and recall of the classifications, combining both attributes.  The NER tasks are evaluated for entity-level using precision, recall, macro $F_1$, and span-level exact matches.  The Macro $F_1$ used here measures this over the multiple attributes of the NER task.  

\subsection{Binary Classification}
Table \ref{tab:binary_metrics} shows the binary classification performance on the 
BBC News and re3d texts. ConfliBERT has high precision for non-conflict texts (a low false positive rate) and high recall for conflict-related texts, suggesting a strong ability to identify relevant content. ConfliBERT's disparity between precision and recall for the conflict class indicates that it flags more texts as conflict-related.  Gemma 2 shows perfect recall for non-conflict texts but fails to identify any conflict-related texts (zero precision, recall, and F1-scores for the conflict class). Meanwhile the Gemma 2 and Llama 3.1, while achieving a comparable overall accuracy, lack the nuanced understanding required for the specific task.  Both have high recall and precision for the non-conflict texts but their performance on the conflict class remains poor and  consequently have lower F1-scores. While Llama 3.1 is marginally better at detecting conflict-related content compared to Gemma 2, it still struggles significantly with this classification task. 

Gemma 2 and Llama 3.1 show a bias towards classifying texts as non-conflict compared to ConfliBERT --- evident from their high precision and recall for the non-conflict class, coupled with poor performance on the conflict class. This imbalance suggests that general LLMs may overfit the majority class (non-conflict), potentially due to class imbalance in the training data or limitations in their ability to capture the nuanced features that distinguish conflict-related texts. While all three models achieve similar overall accuracy, ConfliBERT demonstrates a more balanced and nuanced understanding of both classes, making it more suitable for real-world applications where identifying conflict-related content is crucial. The performance of Gemma 2 and Llama 3.1 shows that for a basic classification task, a domain-specific model that focuses on a local context is likely superior to a larger more general model when put to the same task.  We turn to the issue of further fine-tuning the Llama, Gemma and related models below.

\begin{table}[tp]
\centering
\begin{tabular}{lllrrrr}
\toprule
Model & Class & Precision & Recall & F1-score & Support \\
\midrule
\multirow{4}{*}{ConfliBERT} & 0 (Non-conflict) & 0.9827 & 0.8439 & 0.9080 & 269 \\
& 1 (Conflict) & 0.5385 & 0.9245 & 0.6806 & 53 \\
& Macro Avg & 0.7606 & 0.8842 & 0.7943 & 322 \\
& Weighted Avg & 0.9096 & 0.8571 & 0.8706 & 322 \\
\midrule
\multirow{4}{*}{Gemma 2 (9B)} & 0 (Non-conflict) & 0.8354 & 1.0000 & 0.9103 & 269 \\
& 1 (Conflict) & 0.0000 & 0.0000 & 0.0000 & 53 \\
& Macro Avg & 0.4177 & 0.5000 & 0.4552 & 322 \\
& Weighted Avg & 0.6979 & 0.8354 & 0.7605 & 322 \\
\midrule
\multirow{4}{*}{Llama 3.1 (8B)} & 0 (Non-conflict) & 0.8375 & 0.9963 & 0.9100 & 269 \\
& 1 (Conflict) & 0.5000 & 0.0189 & 0.0364 & 53 \\
& Macro Avg & 0.6688 & 0.5076 & 0.4732 & 322 \\
& Weighted Avg & 0.7819 & 0.8354 & 0.7662 & 322 \\
\bottomrule
\end{tabular}
\caption{Detailed Performance Metrics for Binary Classification}
\label{tab:binary_metrics}
\end{table}

\subsection{Named Entity Recognition Results}

Table \ref{tab:ner_metrics} shows the NER performance of the models for re3d. ConfliBERT performs best across all key metrics with a weighted F1-score of 0.5981, compared to 0.3987 for Gemma 2 and 0.3809 for Llama 3.1.  Noteworthy is ConfliBERT's performance in identifying non-entity tokens (class O) with an F1-score of 0.7493, higher than Gemma 2 (0.5709) and Llama 3.1 (0.5454). ConfliBERT is more robust distinguishing between entity and non-entity tokens, a fundamental task for political conflict text NER.

\begin{table}[tp]
\centering
\begin{tabular}{llrrr}
\toprule
Model & Class & Precision & Recall & F1-score \\
\midrule
\multirow{3}{*}{ConfliBERT} 
& Non-Entity (O) & 0.7728 & 0.7271 & 0.7493 \\
& Macro Avg & 0.1871 & 0.1188 & 0.1407 \\
& Weighted Avg & 0.6510 & 0.5654 & 0.5981 \\
\midrule
\multirow{3}{*}{Gemma 2 (9B)} 
& Non-Entity (O) & 0.7369 & 0.4660 & 0.5709 \\
& Macro Avg & 0.0030 & 0.0019 & 0.0023 \\
& Weighted Avg & 0.5146 & 0.3254 & 0.3987 \\
\midrule
\multirow{3}{*}{Llama 3.1 (8B)} 
& Non-Entity (O) & 0.7335 & 0.4341 & 0.5454 \\
& Macro Avg & 0.0011 & 0.0006 & 0.0008 \\
& Weighted Avg & 0.5123 & 0.3032 & 0.3809 \\
\bottomrule
\end{tabular}
\caption{Detailed Performance Metrics for Named Entity Recognition}
\label{tab:ner_metrics}
\end{table}

The precision and recall values for the non-entity class underscore ConfliBERT's superiority. It maintains the balance between precision (0.7728) and recall (0.7271) compared to its counterparts. Both Gemma 2 and Llama 3.1 show a notable disparity between their precision and recall values, indicating potentially biased NER predictions, despite being larger models compared to ConfliBERT. This suggests that model size alone does not guarantee better performance in specialized NLP tasks like NER for political conflict texts.  ConfliBERT's architecture or training approach may be better suited for this specific task with more appropriate balance in precision and recall.

The macro and weighted average metrics provide insight into the models' performance across all entity classes. ConfliBERT's macro average F1-score (0.1407), while low, is higher than those of Gemma 2 (0.0023) and Llama 3.1 (0.0008). This indicates that ConfliBERT performs more consistently across different entity types, whereas the larger models struggle with certain classes. ConfliBERT demonstrates a clear advantage in this NER task, outperforming larger models across all metrics. Its balanced performance in identifying non-entity tokens and its higher macro average scores suggest a more robust and versatile approach to named entity recognition. These results highlight the importance of specialized domain architectures and training strategies in tackling complex NLP tasks, rather than relying solely on model size.

\subsection{Computational Performance Comparison}
\label{sec:speedeval}

For both the BC task using the BBC News data and the NER task with the re3d, we performed experiments on a machine with GPU capabilities, and recorded the execution time, memory usage, CPU usage, and GPU usage for each LLM.\footnote{These tests were conducted on a machine with an AMD Ryzen 7840HS Processor, NVIDIA GeForce RTX 4060 Laptop GPU 8GB GDDR6 (105W), and 32 GB 6400MHz LPDDR5X RAM. The software environment consisted of WSL (Ubuntu) with Python 3.10.12.}  Table \ref{tab:performance_results} presents the timings for each model and task combination. The most striking difference is the execution time: for classification, ConfliBERT took only 3.52 seconds, while Llama 3.1 (Gemma 2) took 575.23 (730.14) seconds. For NER, ConfliBERT completes the task in 1.42 seconds, compared to 489 (866) seconds for Llama 3.1 (Gemma 2).  The speed of ConfliBERT can be attributed to its parallel architecture for processing input data.  ConfliBERT's superior performance stems from its ability to process inputs in parallel. BERT-based models, like ConfliBERT, can efficiently batch multiple inputs and process them simultaneously.  Generative LLMs like Gemma and Llama, typically process inputs sequentially so each text requires a separate request to the model, introducing additional computational overhead. While we parallelize these models by batching multiple task requests, there are context-length constraints on processing the texts that differ across the models.

\begin{table}[tp]
\centering
\begin{tabular}{llrrr}
\toprule
Model & Task & Execution Time (s) & Max Memory (MB) & GPU Usage (\%) \\
\midrule
\multirow{2}{*}{ConfliBERT} 
& Classification & 3.52 & 919.94 & 95.93 \\
& NER & 1.42 & 950.63 & 95.63 \\
\midrule
\multirow{2}{*}{Llama 3.1 (8B)} 
& Classification & 575.23 & 950.64 & 92 - 94 \\
& NER & 489.39 & 950.65 & 92 - 94 \\
\midrule
\multirow{2}{*}{Gemma 2 (9B)} 
& Classification & 730.14 & 950.65 & 90 - 97 \\
& NER & 866.23 & 957.19 & 90 - 97 \\
\bottomrule
\end{tabular}
\caption{Performance Metrics for ConfliBERT, Llama 3.1, and Gemma 2 Models}
\label{tab:performance_results}
\end{table}

\section{Classifying Texts about Terrorist Attacks}

Some pre-training of the generative LLMs could bring their performance up to or exceeding the performance of ConfliBERT \citep[see,][commenting on \citet{haffner2023introducing}]{Wang_2023}.  Fine-tuning the LLMs like ConfliBERT permits an appropriate and critical comparison since  learning additional parameters of neural layers in the BERT representation of the texts in them is critical for additional extensions of models like ConfliBERT.  This asks for a comparison of ConfliBERT to other more recent generative LLMs \emph{with pre-training on political conflict texts}.  Critical to this example is its replicability and service as a baseline comparison for event feature classification across new LLMs.  
This example replicates the common researcher problem: one has identified political conflict-related texts (or prior dataset to be extended) and organized them (say in a CSV, JSON, or other database) for analysis with standard NLP to extract the relevant event information.  This then leaves open the choices of the LLM and the pre-training.  As an illustration, consider the short texts in the Global Terrorism Dataset (GTD) \citep{lafree2007introducing}.\footnote{One consideration in selecting this example is the need for shareable and non-copyrighted texts for replication. This removes a barrier to entry for replication that would exist if data requiring a copyright, extensive downloads (scraping) or a license is used -- e.g., Linguistic Data Consortium corpora.  Further, we want something that is timely and relevant to conflict scholars.  The short descriptions from GTD described here fit the bill.}  GTD is a good choice because it 1) is a comprehensive open-source database of terrorist events, 2) contains the conflict classification tasks (what kind of attack is in the event?), 3) provides consistent, well-structured texts for NLP tasks, and 4) is classified by experts: one knows from the codebook and the dataset who perpetrated the terrorist attacks, the nature of the attacks and the types of victims.  One cannot use these texts for the BC task, but they are suitable for evaluating models' NER and event multi-label classifications.


The task here is predicting the categorization of terrorist attacks from each GTD text description, comparing the various LLMs' codings to the original (human) GTD annotations of the terrorist attack types.  ConfliBERT is compared to the aforementioned Llama and Gemma varieties, a larger LLM (Qwen 2.5),  and a fine-tuned variant of Llama that we denote as ConflLlama.\footnote{ConflLlama is a base Llama model given the training data and the classification prompt in the Appendix.}  The training prompts for the generative LLMs are given in an appendix.  
The selection of evaluation metrics (ROC, accuracy, precision, recall, and F1-score) follows standard practices in conflict event classification \citep{schrodt2020automated}. 

For testing and evaluation GTD data from 1970 to 2016 are used to train the LLMs and they are tested with data from 2017--2020.  Most of the GTD events have no texts for 1970-1997, so this is mainly based on training texts from 1998--2016.  The LLM coded texts produce sets of binary classifications of each of the nine GTD event types across 37,709 texts recorded in GTD using each of the six models (ConfliBERT, ConflLlama4, ConflLlama8, Gemma, Llama, and Qwen).  The first three of these are ones we produce, the latter three are "off the shelf" from \url{huggingface.co}. 

\subsection{Basic Classification Results}

\begin{figure}[t]
    \centering
    \includegraphics[trim = 0 0 0 30,width=0.95\linewidth]{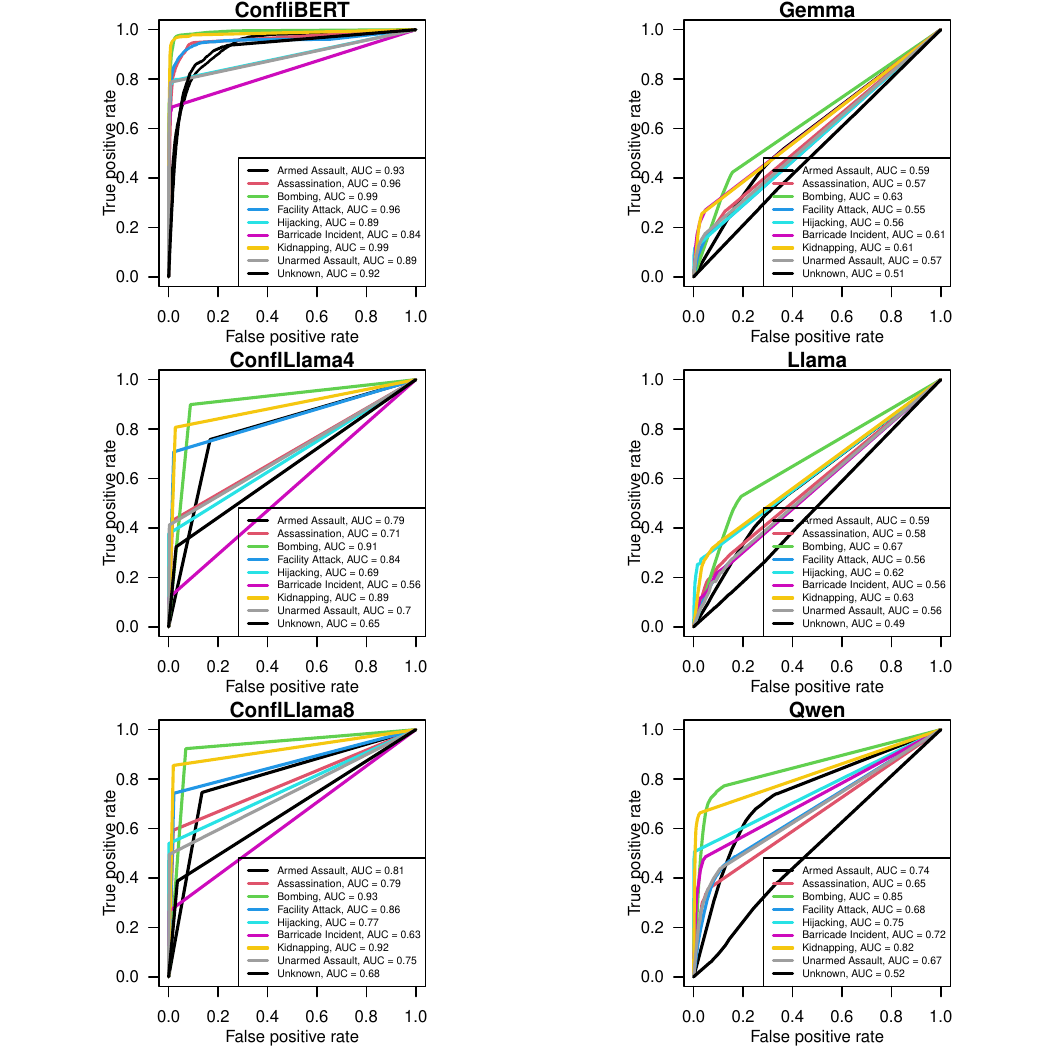}
    \caption{ROC and AUC for each LLM and event type.  Curves along the northwestern edge are better.}
    \label{fig:GTD-ROC}
\end{figure}

Figure \ref{fig:GTD-ROC} shows the comparative analysis of model performance differences across the LLM architectures. Here the left column presents receiver-operator characteristic curves (ROCs) for the conflict-trained LLMs while the right column presents the same for the general (non-conflict data trained), generative LLMs.  The results in the right column are closer to a $45^{\circ}$ line, indicating nearly random classification performance by event-type.  The area under the curve (AUC) for each event type are in the lower right.\footnote{AUC scores are relevant in conflict event classification due to balance precision and recall \citep{schrodt2020automated}}. Across models, the higher accuracy of the ConfliBERT is evident and generally best for events about bombings and kidnappings (the green and gold lines) across the models -- the most common kinds of attacks.  

\begin{figure}[tp]
    \centering
    \includegraphics[trim = 0 0 0 60, width=0.95\linewidth]{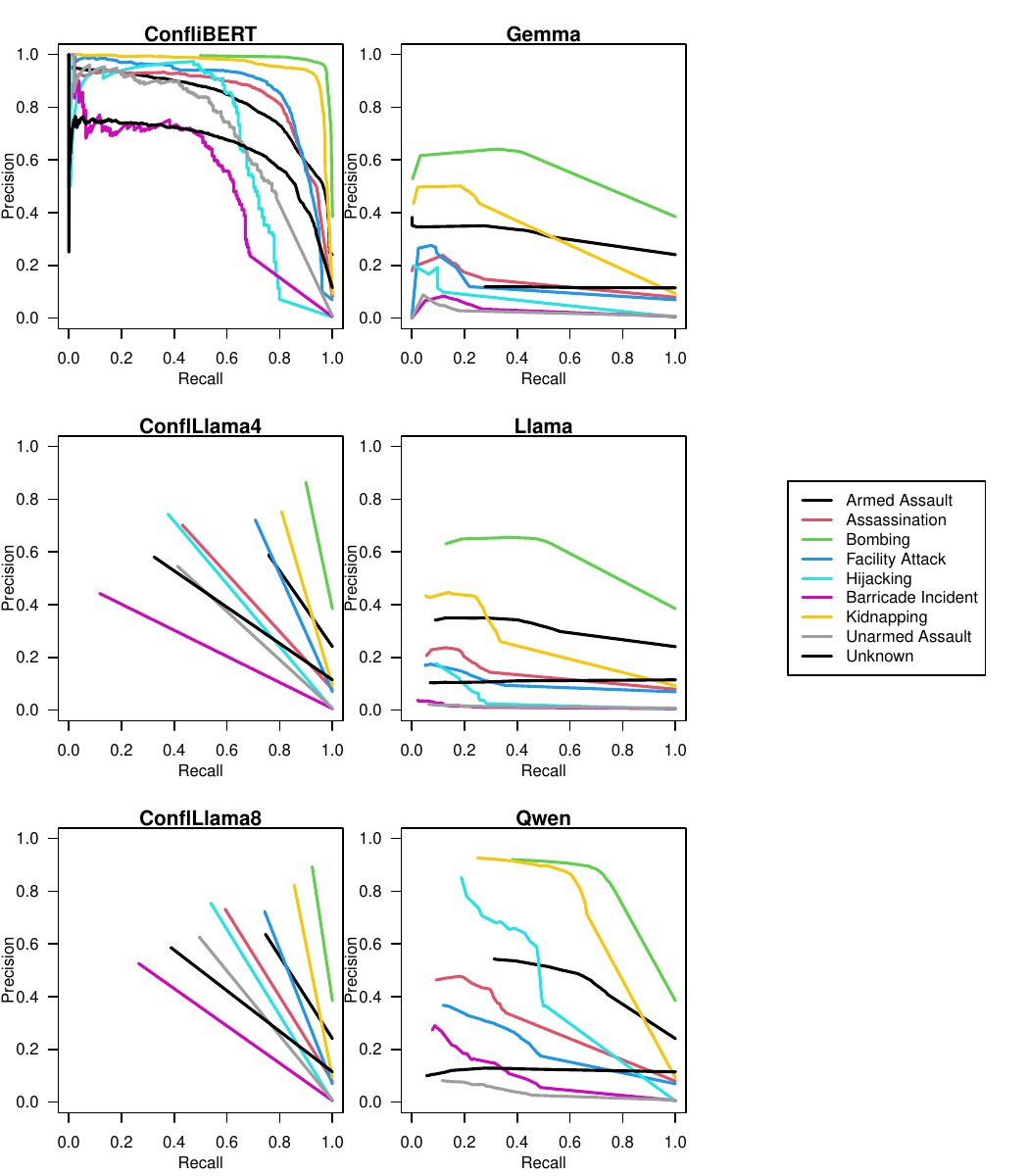}
    \caption{Precision-recall Curves for each LLM and event type.  Curves along the northeastern edge are better.}
    \label{fig:PR-curves}
\end{figure}

One criticism of only using an accuracy to compare models is that it is inflated by predicting the dominant class for imbalanced problems like the classifications here.   Figure \ref{fig:PR-curves} shows the models' precision-recall curves, in parallel with Figure \ref{fig:GTD-ROC}.  Best precision-recall curves are those that follow the top, northeastern edge of the plot.\footnote{The numeric precision and recall scores commonly see in tables are weighted averages over the appropriate axes of these plots.} ConfliBERT has the highest precision-recall combinations for similar events (i.e., for the same colored lines).  Of the larger generative AI models, only the more recent and much larger Qwen model comes close to ConfliBERT and ConflLlama in precision and recall performance, but only for kidnappings and bombings.

\begin{figure}[tp]
    \centering
    \includegraphics[trim = 0 0 0 50, width=0.95\linewidth]{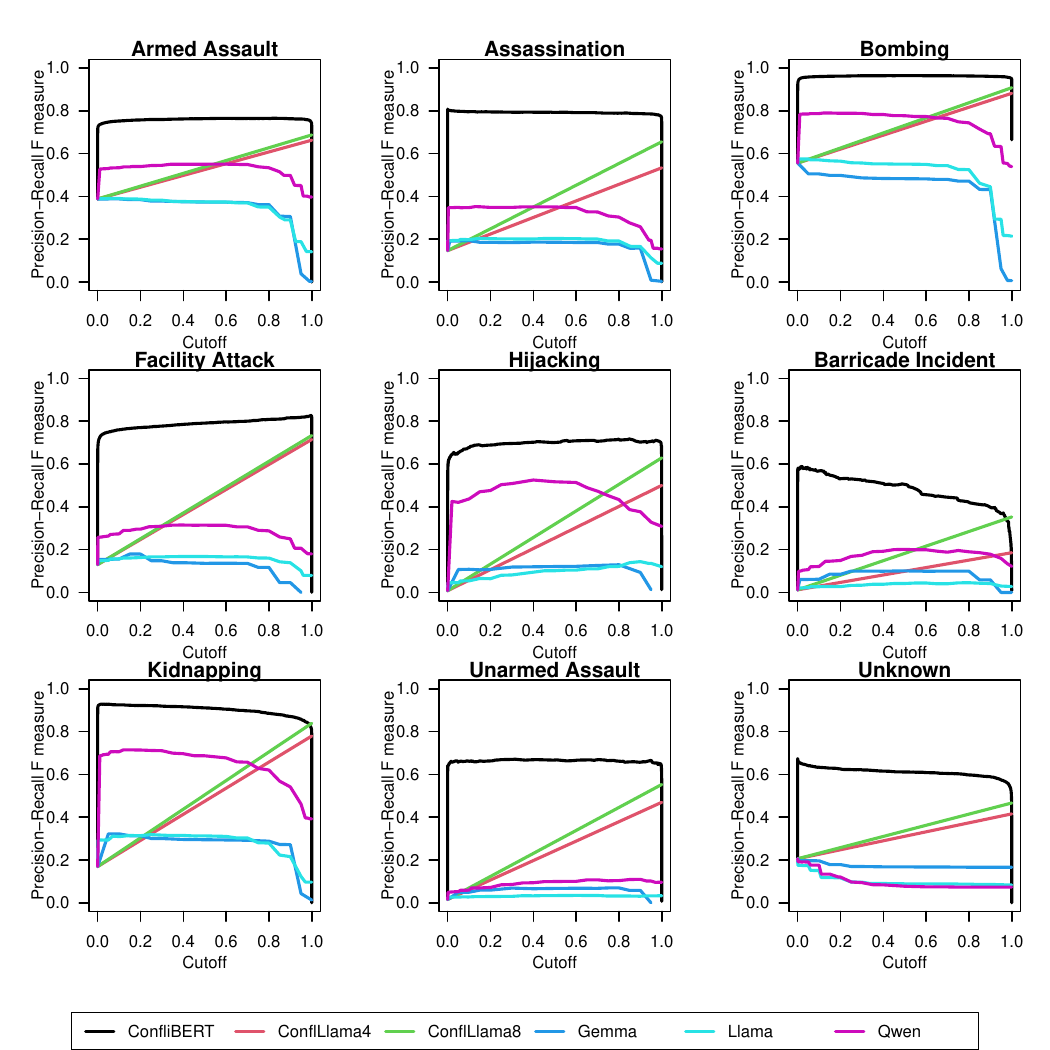}
    \caption{F-scores across cutoffs for each event type model.  Higher curves are better.}
    \label{fig:F-curves}
\end{figure}

Precision-recall curves are a function of the cutoff used to classify a prediction as a match to GTD.  Choosing the wrong cutoff, one may miss the benefits of a model to detect events (and mis-state their precision and recall in Figure \ref{fig:PR-curves}).  Figure \ref{fig:F-curves} presents the $F$ score for the precision-recall as a function of the chosen cutpoint for the correct classification \emph{for each event type}.  Unlike Figures \ref{fig:GTD-ROC} and \ref{fig:PR-curves}, these are grouped by types of events, so the colors used indicate the models here.  Ideally the values should be high across the cutpoints, like those for ConfliBERT. ConfliBERT and Qwen have the best $F$ scores, followed by ConflLlama.  These results align with previous findings suggesting that domain-specific fine-tuning often outperforms larger, general purpose models \citep{gururangan2020don}. 
Like in other specialized domains, ConfliBERT's strong performance can be attributed to its training on conflict-related data.    

For the GTD conflict-related text analysis tasks, ConfliBERT out performs the baseline competitors across all metrics shown in aggregate in Table \ref{tab:model_performance}.  Its considerable speed improvements over larger models also reflect broader trends in NLP research emphasizing the importance of computational efficiency \citep{schwartz2020green}.\footnote{Processing times were measured on identical hardware configurations to ensure fair comparison.} 
For the fewer than 40K sentences evaluated here this is remarkably fast, yet for a larger document processing-training problem, the more general generative LLMs like Gemma, Llama, and Qwen are likely computationally prohibitive.  While general-purpose large language models continue to improve, these results reinforce previous findings that specialized models can achieve superior performance in domain-specific tasks while maintaining significantly lower computational requirements \citep{strubell2019energy}.

\begin{table}[t!]
\centering
\begin{tabular}{l|rrrr|rrr}
\toprule
Model & Accuracy & Precision & Recall & F1 &  Total & Time / & Relative \\ 
& & & & & Time & Document & Speed\\
\midrule
ConfliBERT & 0.90 & 0.83 & 0.77 & 0.79 & 27.6s & 0.0016s & 759.49x\\
ConflLlama-Q4KM$^*$ (8B) & 0.72 & 0.72 & 0.72 & 0.71 & 49.9m & 0.1746s & 7.15x\\ 
ConflLlama-Q8$^*$ (8B) & 0.76 & 0.76 & 0.76 & 0.75 & 52.3m & 0.1831s & 6.82x\\
Gemma 2 (9B) & 0.60 & 0.27 & 0.19 & 0.21 & 3.1h & 0.6605s & 1.89x\\
Llama 3.1 (8B) & 0.52 & 0.13 & 0.12 & 0.11 & 3.3h & 0.7191s & 1.74x\\
Qwen 2.5 (14B) & 0.74 & 0.50 & 0.44 & 0.45 &  5.8h & 1.2490s & 1.00x \\
\bottomrule
\end{tabular}
\caption{Model Performance Comparison (Macro Averages). $^*$ConflLlama timing measurements were performed on Delta HPC resources and are not directly comparable to other models' timing metrics.}
\label{tab:model_performance}
\end{table}

\subsection{Multi-Label Classification Performance}

Incidents that involve more than one event type are documented with multi-label classifications in the GTD. This occurs say when an incident includes an armed attack or assault in the course of a kidnapping.  Multi-label classification is important in conflict event coding, as real-world events often exhibit characteristics of multiple attack types \citep{radford2021automated}.  Less than 10\% of the post-2016 (the test period) data has multi-label events. Originally, ConfliBERT was not built to handle this task, but can be used to accommodate it.   Multi-label classification results, presented in Table \ref{tab:multilabel_metrics}, demonstrate ConfliBERT's ability to handle complex event categorizations. The model achieved a subset accuracy of 79.38\% and the lowest Hamming loss (0.035), indicating superior performance in scenarios where events may belong to multiple categories.  The close alignment between predicted label cardinality (0.907) and true label cardinality (0.963) suggests that the model has effectively learned to capture the complexity of conflict events without over- or under-predicting multiple classifications.

\begin{table}[tp]
\centering
\begin{tabular}{lrrrrrr}
\toprule
Metric & ConfliBERT & ConflLlama-Q8 & ConflLlama-Q4 & Qwen & Gemma & LLaMA \\
\midrule
Subset Accuracy (\%) & 79.38 & 72.40 & 68.80 & 50.99 & 30.70 & 32.03 \\
Hamming Loss & 0.035 & 0.052 & 0.061 & 0.096 & 0.133 & 0.148 \\
Partial Match (\%) & 79.66 & 73.80 & 71.10 & 55.04 & 30.65 & 35.64 \\
\midrule
\multicolumn{7}{l}{Label Cardinality} \\
True & 0.963 & 0.963 & 0.963 & 0.963 & 0.963 & 0.963 \\
Predicted & 0.907 & 0.975 & -- & 0.903 & 0.711 & 0.932 \\
\bottomrule
\end{tabular}
\caption{Multi-Label Classification Metrics}
\label{tab:multilabel_metrics}
\end{table}

The performance of ConfliBERT across all metrics suggests several important implications for conflict event classification. First, the results demonstrate that ConfliBERT with domain-specific fine-tuning can substantially outperform larger, general-purpose models, even when the latter have significantly more parameters \citep{gururangan2020don}. The model's strong performance on rare event types is particularly noteworthy, as it addresses a common challenge in conflict event classification \citep{mueller2021forecasting}. This suggests that the fine-tuning process successfully captures the nuanced characteristics of different attack types, even with limited training examples.

\subsection{Validity Comparisons}

Another assessment of the classification differences from the LLMs is to consider how their distributions change over the event types.  At any one point in (recent) time it may not be evident how the (mis-) classification of a given type of events affects inferences.  But if there were systemic biases in LLM classification they are more evident as more types over events are collected --- an inherently time series process for these data.  This is particularly relevant in say a changepoint analysis of the drivers of trans-national terrorism like that addressed in  \citet[Figures 1--3]{santifort2013terrorist}, who use cumulative sums of terrorist event type classifications over time that would be severely biased upward or downward by LLM mis-classifications like those documented above.

\begin{figure}[tp]
    \centering
    \includegraphics[trim = 0 0 0 50, width=0.95\linewidth]{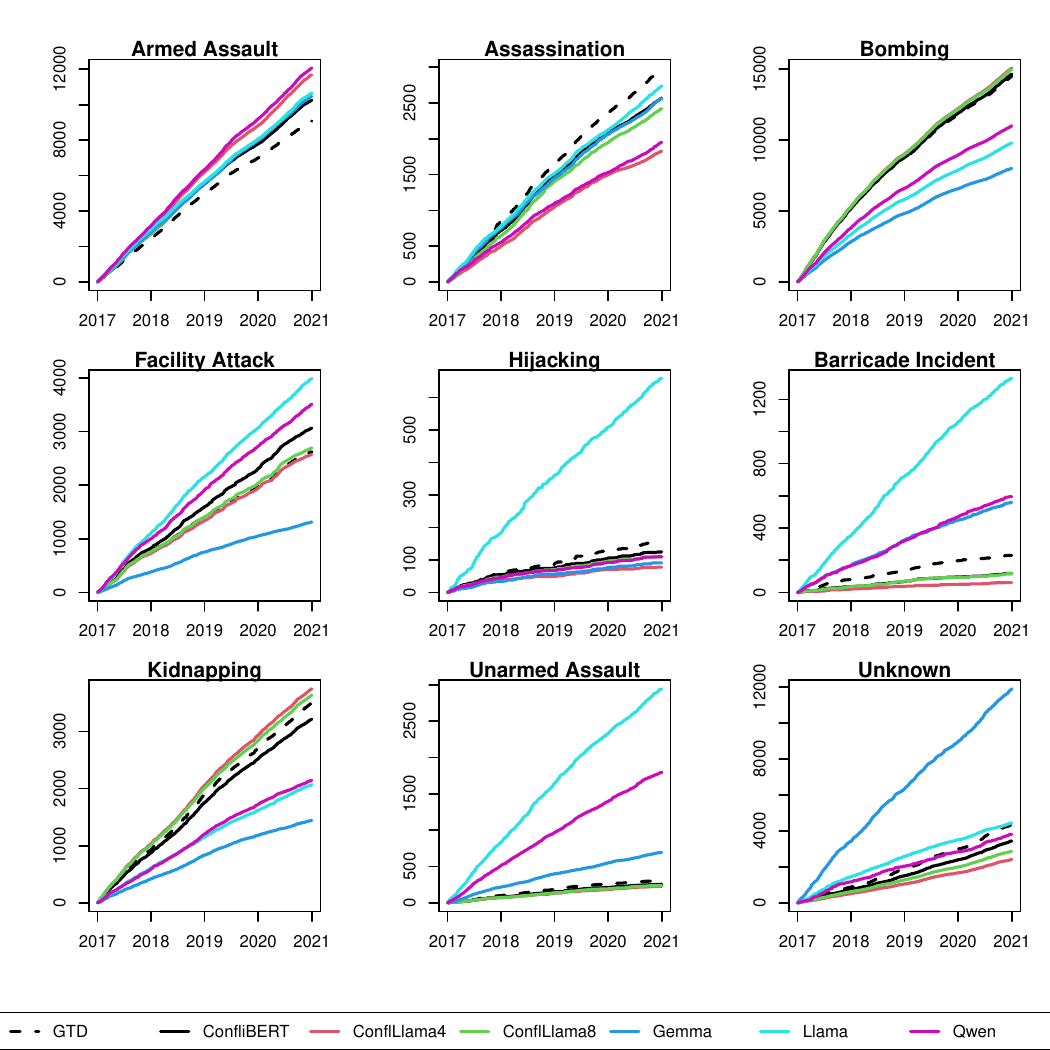}
    \caption{Cumulative number of predicted events, 2017--2021 by type and model}
    \label{fig:GTD-cumts}
\end{figure}

Figure \ref{fig:GTD-cumts} shows the cumulative time series of number of each type of GTD terrorist event from 2017--2020 as a dashed line.  LLMs whose classifications are above this line are over-predicting / over-classifying the number of events of a given type while those under the dashed line are the reverse.  A few immediate patterns jump out: the non-conflict pre-trained LLMs under classify bombing events (uppermost-right plot) -- so Gemma, Llama, and Qwen.  Second, the Llama, Qwen and Gemma models generally do poorly with the rarest event types (hijackings, barricade incidents, and unarmed assaults), but their performance includes over and under predictions relative to GTD's human-coded data.  

The reason such deviations and the relative performance of the LLMs matters here is that it will effect downstream time series analyses since systemtically mis-measured event types will lead to incorrect time series dynamics and inferences that would confound those in works like \citet{santifort2013terrorist}.  This builds on a key point since it shows that even using more data and more sophisticated methods for encoding texts, the issues of aggregation over time will still be important and affect inferences \citep{Shellman_2004}.

\section{Conclusions and Next Steps}

Conflict research and event data have a fruitful history of incorporating NLP methods to advance methods of unstructured data processing (\citet{schrodt2001automated,schrodt2012precedents,beieler2016generating}) and this work is a continuation of that path.  The adoption of NLP techniques like those employed by ConfliBERT improves how political scientists can extract and study events and political interactions. These tools offer the potential to analyze larger volumes of text data, enabling more comprehensive studies. They can reduce bias in event coding by applying consistent criteria across large datasets, identify patterns and trends that might be missed by human coders, and enable near-real-time analysis of political events as they unfold. While LLMs generally hold this potential, the approach in ConfliBERT uses domain-specific knowledge, resulting in superior performance and even faster data processing of text classification and summarization tasks.   

There are a series of conclusions to be drawn from this analysis. The results leverage existing infrastructure of BERT-alike LLMs and conflict researchers’ expertise to advance scholarship on conflict processes and international security.  The contribution is that domain-specific knowledge -- the things international and civil conflict scholars know -- should be part of the information extraction process used to 1) filter relevant reports (binary classification) and 2) identify events, and 2) annotate their attributes (named entity recognition).   A BERT-based model \emph{plus} domain knowledge is able to do this in a way that is better on several metrics as documented in Section \ref{sec:eval}.

Beyond an infrastructure outline for political scientist to engage with texts about conflict and violence there are several other contributions of note here.  First, ConfliBERT builds on a known ontology (CAMEO/PLOVER) \citep{schrodt2012precedents} for coding events and provides a set of tools for continuing to do so.  This allows for additional fine-tuning of the models and a flatter development and learning curve.   Unlike current large scale general LLMs, this allows researchers to openly and quickly work in this area (the span from ConfliBERT in \citet{hu2022conflibert} to the recent paper \citet{osorio_2024} is less than 36 months.)  

Second, the typical social science conflict researcher need not build their own ConfliBERT: one can fine-tune or extend this model since it is open and available for use via our website and HuggingFace \url{https://huggingface.co/eventdata-utd}. About 200GB of combined training data are invested in ConfliBERT, ConfliBERT Spanish, and ConfliBERT Arabic.  
Additional classifications and training based on new ideas, texts, actors, etc. can be added and evaluated.  We have done in this in the efforts to extend beyond event coding just in English by working not just with a language and domain specific dictionary approach \citep{osorio2017supervised}, but a general BERT-like model in Spanish \citep{yang2023conflibert}.  This shows how the domain-specific approaches can be brought to bear from say old codebooks and ontologies into the LLMs \citep{hu-etal-2024-leveraging}.  So prior domain knowledge about regions, languages, and events can be part of how LLMs are used to encode and understand new texts and data.  

Third, this approach is sometimes better than using larger LLMs.  Unlike large generative LLMs, a decoder model like ConfliBERT better fits what a social scientist needs, which is data extraction, organization, and (predictive) classification.  Section \ref{sec:eval}  shows this in terms of performance metrics like accuracy, precision, $F_1$, etc.   It is also faster to use ConfliBERT.   While the initial LLM training for ConfliBERT and its language variants took thousands of GPU hours, the work in Section \ref{sec:speedeval} only takes hours of computing time on current laptops.  Deploying this on a real data problem is scalable and feasible: it is 300-400 times faster than using a proprietary LLM for NER and 150-200 times faster for binary classification.

Finally, there are problems that can be addressed such as learning about and connecting events and actors.  One area of interest is extending ontologies and NER to recognize and learn about new events and actors -- who is the next leader, insurgent, or what are they doing?  This is related to a literature on continual learning and catastrophic forgetting in LLMs.  There is work in this area that can be applied and used to aid models like ConfliBERT as well \citep[e.g.,][]{li-etal-2022-lpc}.  This would also be useful for extending text-as-data methods across networks of texts, languages, etc.

\newpage

\appendix


\section{Appendix: LLM prompts}

ConfliBERT and ConflLlama are fine-tuned specifically for terrorist event classification, without explicit prompting for output classifications given input event texts. For the general-purpose large language models (e.g., Gemma, Qwen, and Llama) the following structured prompt is used:

\begin{example}
\textcolor{blue}{Prompt:} "Classify each of the following events into up to three of these categories, providing probabilities for each:
Assassination, Armed Assault, Bombing/Explosion, Hijacking,
Hostage Taking (Barricade Incident), Hostage Taking (Kidnapping),
Facility/Infrastructure Attack, Unarmed Assault, Unknown

For each event, return only a JSON object with category names as keys and probabilities as values.
Example format: \{"Armed Assault": 0.7, "Bombing/Explosion": 0.2, "Unknown": 0.1\}

Events:"
\end{example}

\noindent This prompt follows key principles on effective LLM prompting \cite{liu2023pre, wei2022emergent}. Its structured format with explicit probability requirements builds on research showing that quantitative outputs improve model classification tasks \citep{brown2020language}. The multi-label approach, allowing up to three categories, reflects the complex classification task and the the original GTD structure -- allowing direct comparisons. The JSON output format facilitates consistent parsing and evaluation, addressing challenges in systematic event coding. This standardization enables direct comparison with both human annotations and across models, while maintaining interpretability.

\newpage
\singlespacing
\bibliography{apsa-bib}

\end{document}